%% file: acl2023.tex
\pdfoutput=1

\documentclass[11pt]{article}

\usepackage[]{ACL2023}

\usepackage{times}
\usepackage{latexsym}
\usepackage{booktabs}
\usepackage{graphicx}
\usepackage{multicol}
\usepackage{multirow}
\usepackage{makecell}

\usepackage{amsmath}
\usepackage{amsfonts}
\usepackage{amssymb}

\usepackage[T1]{fontenc}

\usepackage[utf8]{inputenc}

\usepackage{microtype}

\usepackage{inconsolata}

\usepackage{CJK}
\usepackage{longtable}
\usepackage{supertabular}
\usepackage{colortbl}
\definecolor{amaranth}{rgb}{0.9, 0.17, 0.31}
\definecolor{highlightgray}{RGB}{68,68,68}
\definecolor{highlightgreen}{RGB}{72,142,35}
\definecolor{highlightblue}{RGB}{8,67,138}
\definecolor{highlightred}{RGB}{140,15,13}

\title{SeqGPT: An Out-of-the-box Large Language Model for Open Domain Sequence Understanding }

\author{
Tianyu Yu$^1$\thanks{~~Equal first authorship.}, Chengyue Jiang$^2$\footnotemark[1], Chao Lou$^2$\footnotemark[1], Shen Huang$^4$\footnotemark[1], Xiaobin Wang$^4$ \\
\textbf{Wei Liu$^2$, Jiong Cai$^2$, Yangning Li$^1$, Yinghui Li$^1$, Kewei Tu$^2$, Hai-Tao Zheng$^1$}\\\textbf{Ningyu Zhang$^3$, Pengjun Xie$^4$, Fei Huang$^4$, Yong Jiang$^4$}\thanks{~~Corresponding author. \\ This work was conducted when Tianyu Yu, Chengyue Jiang, Chao Lou, Wei Liu, Jiong Cai,  Yangning Li and Yinghui Li were interning at Alibaba DAMO Academy.}\\
$^1$Tsinghua University\qquad $^2$ShanghaiTech University \\
$^3$Zhejiang University \qquad $^4$DAMO Academy, Alibaba Group\\
\texttt{yiranytianyu@gmail.com}, 
\texttt{\{jiangchy,louchao\}@shanghaitech.edu.cn} \\
\texttt{\{pangda,xuanjie.wxb,yongjiang.jy\}@alibaba-inc.com} \\
}

\begin{document}
\maketitle
\begin{abstract}

Large language models~(LLMs) have shown impressive ability for open-domain NLP tasks. However, LLMs are sometimes too footloose for natural language understanding~(NLU) tasks which always have restricted output and input format. Their performances on NLU tasks are highly related to prompts or demonstrations and are shown to be poor at performing several representative NLU tasks, such as event extraction and entity typing. To this end, we present SeqGPT, a bilingual (i.e., English and Chinese) open-source autoregressive model specially enhanced for open-domain natural language understanding. We express all NLU tasks with two atomic tasks, which define fixed instructions to restrict the input and output format but still ``open'' for arbitrarily varied label sets. The model is first instruction-tuned with extremely fine-grained labeled data synthesized by ChatGPT and then further fine-tuned by 233 different atomic tasks from 152 datasets across various domains. The experimental results show that SeqGPT has decent classification and extraction ability, and is capable of performing language understanding tasks on unseen domains. We also conduct empirical studies on the scaling of data and model size as well as on the transfer across tasks. Our model is accessible at \url{https://github.com/Alibaba-NLP/SeqGPT}.
\end{abstract}

\section{Introduction}

\begin{figure}[tb]
    \centering
    \includegraphics[width=0.86\columnwidth]{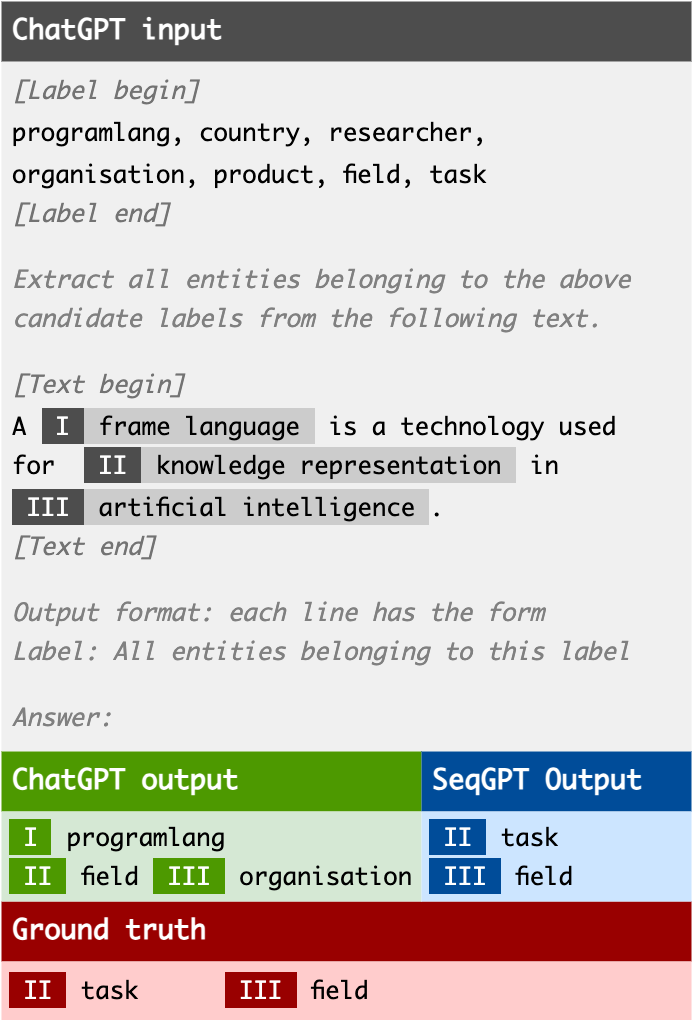}
    \caption{An example of ChatGPT and SeqGPT performing the CrossNER task in the zero-shot setting. ChatGPT mislabeled entities, while SeqGPT succeeded. \textit{\textcolor{gray}{Italic gray texts}} are the prompt template. SeqGPT uses a different prompt, as shown in Figure~\ref{fig:overview}.}
    \label{fig:intro_demo}
\end{figure}

Recent advancements in large language models (LLMs) have demonstrated their impressive ability across various NLP tasks~\cite{kaplan2020scaling,wei2022emergent,chung2022scaling,zhao2023survey,DBLP:journals/corr/abs-2307-09007}.
Regarding natural language understanding (NLU) tasks, although the next-word-prediction approach utilized by language models implies little bias to the task-specific output structures, such as spans in named entity recognition~(NER) and triplets in relation extraction~(RE), numerous attempts~\cite{qin2023chatgpt,wei2023zeroshot,wadhwa-etal-2023-revisiting,ashok2023promptner} have been made to apply LLMs to open-domain NLU tasks through the application of prompt engineering, mainly due to the LLMs' exceptional ability of generalization and instruction-following (Figure~\ref{fig:intro_demo}).
However, the direct application of LLMs comes with notable drawbacks. Instruction-following necessitates the use of a sufficiently large model~\cite{kaplan2020scaling,wei2022emergent}, for example, GPT-3~\cite{brown2020language} has 175B parameters, which can lead to considerable inference costs and challenges in customization~\cite{hu2022lora,liu2022fewshot,liu-etal-2022-p}.
In addition, prompt engineering is crucial to achieve promising performance and ensure adherence to output format standards. However, it is highly empirical and the models may not consistently abide by it~\cite{chase_langchain_2022,autogpt}.

\begin{figure*}[tb]
    \centering
    \includegraphics[width=0.95\textwidth]{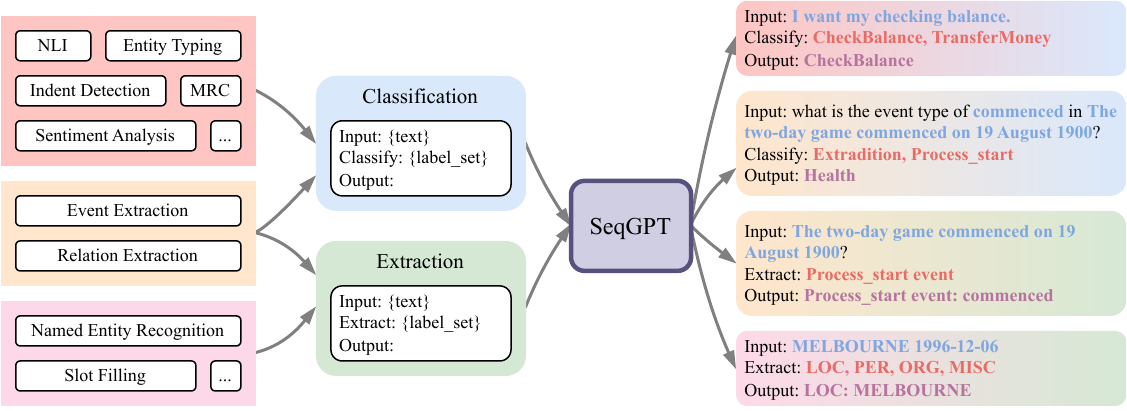}
    \caption{The overview of SeqGPT. Each NLU task is translated into atomic tasks with consistent input-output formats. Black/blue/red/purple tokens are templates/inputs/query or label lists/outputs.}
    \label{fig:overview}
\end{figure*}
To perform NLU tasks more effectively, some researchers~\cite{wang-etal-2022-deepstruct,wang2023instructuie,lu2023pivoine,10.1145/3485447.3511998,deepke-llm} have focused on continuing to train moderate-sized foundation models (approximately 10B parameters, e.g., BLOOM-7B1~\cite{workshop2023bloom}),
which not only improve computational friendliness but also deliver competitive capabilities, in a manner of unifying various tasks. Data consumed in the training procedure can be sourced from either an aggregation of existing close-domain datasets~\cite{wang-etal-2022-deepstruct,wang2023instructuie} or open-domain but noisy datasets generated through approaches such as weak supervision~\cite{lu2023pivoine} and interaction with LLMs~\cite{wang-etal-2023-self-instruct}.
The extra training purportedly empowers moderate-sized models to surpass their large-scale counterparts in zero-shot performance across various NLU benchmarks.
These tuned models can also provide a stable standard output interface, making evaluation and downstream application convenient.

Our research is in the line of enhancing the NLU ability of LLMs via training but involves a broader range of NLU tasks and incorporates a greater diversity of open-domain data than previous work. This is motivated by recent instruction tuning studies, which emphasize the advantages of enhancing task diversity rather than simply increasing data volume~\cite{wang-etal-2022-super,iyer2023optiml}. Specifically, we collect and unify 152 datasets across 11 NLU tasks, encompassing not only commonly included information extraction~(IE) tasks like NER~\cite{wang-etal-2022-deepstruct,wang2023instructuie}, but also tasks overlooked in prior work, such as natural language inference~(NLI) and extraction-based machine reading comprehension~(MRC). Moreover, to bridge the discrepancy between practical scenarios and existing close-domain NLU data, we generate a large-scale open-domain dataset from various sources. In contrast to earlier studies on automatic NLU data generation, which typically rely on a single domain source (e.g., Wikipedia) and assign labels based on a predefined knowledge base~\cite{lu2023pivoine}, we instruct ChatGPT to invent appropriate labels for each sample and identify corresponding answers because ChatGPT is proficient at summarizing and producing annotations at a human level~\cite{brown2020language,gilardi2023chatgpt,zhu2023chatgpt}. The generated dataset contains more than 800 thousand distinct reasonable labels, which is substantially richer than previous datasets but remains high quality upon our manual inspection.

Using the two datasets, we train \textbf{Seq}uence understanding enhanced \textbf{GPT}, shortly SeqGPT, based on BLOOMZ~\cite{muennighoff2023crosslingual}, a family of instruction-tuned language models. Our training procedure consists of two stages: initially, pre-training using the diverse, albeit noisy, ChatGPT-generated data and subsequently fine-tuning with the collection of real NLU datasets. This strategy is driven by the intention to first enhance the ability of generalization through the use of diverse data and then refine the model to align with human preferences.
Our experiments revealed that SeqGPT consistently surpasses ChatGPT on zero-shot NLU benchmarks by a large margin. The key findings derived from our study can be summarized as follows:
\begin{itemize}
    \item Scaling up the model size enhances performance.
    \item However, simply scaling up the data size without considering diversity does not consistently yield performance improvements.
    \item Increasing task diversity improves performance, although this increase is logarithmic with respect to the number of tasks.
    \item Larger models are capable of generalizing across languages and tasks.
\end{itemize}






\section{Method}

\subsection{Unified Approach}

In order to solve a novel open-domain task, a language model expects a sequential input encoding both the sentence and necessary knowledge of the task and outputs answers accordingly.
To tackle different NLU tasks with a single model and a consistent input-output format, we consider a unified approach that translates them into two atomic tasks:
\begin{itemize}
    \item \textbf{Extraction (EXT):} This task identifies all relevant spans for each query. A query can be a single word, a phrase (as in traditional extraction tasks), or a natural language description (as in machine reading comprehension and instruction following). 
    \item \textbf{Classification (CLS):} This task aims to associate the entire input with a suitable subset of the given labels, which permits both multi-class and multi-label classification.
\end{itemize}

For each atomic task, we design a simple prompt template, which consists of (1) some control tokens indicating different parts of inputs, (2) the specific text to be analyzed, and (3) a list of queries or labels of interest. Regarding the output, the answers are formatted into fixed and easy-to-parse forms depending on the type of atomic tasks. Particularly, for the extraction task, the answer is listed line by line. Each line contains a user-typed query, followed by a list of phrases as its corresponding answer. We do not require the models to provide the positions from which these phrases are extracted, as transformer-based models are not proficient in token counting. For the classification task, the answer is formatted as a single-line list containing answer labels taken from the provided label set.

Typically, most tasks only involve one of these atomic tasks. NLI and NER exemplify tasks that rely solely on classification or extraction. However, some tasks require decomposition into multiple atomic tasks. For example, relation extraction (RE) is performed first to identify spans, followed by classification to discern the relationships between each span pair. Besides, we make necessary efforts of prompt designing to handle task-specific input. For example, NLI involves two sentences (i.e., premise and hypothesis). We concatenate them with a separator. Figure~\ref{fig:overview} shows a brief illustration, and Section~\ref{sec:tasks_datasets} presents more details. 

Contrary to previous studies on instruction tuning that require significant effort to design task descriptions~\cite{wang-etal-2022-super,wang-etal-2023-self-instruct,wang2023instructuie}, we inject task-specific information to our models via informative queries or labels. Therefore, the model can be generalized to new tasks and domains without human effort to craft new elaborate task descriptions. 
While this approach may potentially limit the performance due to the inflexible prior knowledge injection at inference time, our experiments show that, after continuous training on massive NLU tasks, the model learns how to solve NLU tasks and how to generalize, eliminating the need for additional information in the inference time, such that achieves a balance between efficiency and effectiveness.

As prompts are pivotal to achieving high performance, we examine various design possibilities, such as using language-specific or language-agnostic templates. A thorough discussion and experimental comparison will be in Section~\ref{sec:add_res}.

\begin{table}[tb]
    \centering
    \resizebox{\linewidth}{!}{
    \begin{tabular}{cc|ccc}\toprule
         \textbf{Lang.} & \textbf{Task}& \textbf{\# inst.} & \textbf{\# token} & \textbf{\# label} \\\midrule
         \multirow{ 3}{*}{En} & CLS & 50,172 & 4,914,471 & 22,002\\
         & ET & 212,734 & 21,594,057 & 84,461 \\
         & NER & 60,094 & 9,803,353 & 117,300 \\\midrule
         \multirow{ 3}{*}{Zh} & CLS & 49,917 & 7,283,509 & 32,209 \\
         & ET & 576,839 & 170,318,622 & 143,935 \\
         & NER & 196,515 & 46,210,373 & 417,168 \\\midrule
         \multicolumn{2}{c|}{All} & 1,146,271 & 260,124,385 & 817,075 \\\bottomrule
    \end{tabular}
    }
    \caption{Statistics of the pre-training data.}
    \label{tab:pt_stat}
\end{table}

\subsection{Pre-training Data}

Motivated by recent evidence that scaling data diversity benefits models' generalization ability on unseen data~\cite{wang-etal-2022-super,iyer2023optiml},
we construct a large-scale pre-training (PT) dataset with an extremely diverse label set and multiple source domains, including Wikipedia, news, and medicine. For covering both atomic tasks, we consider three tasks: classification, entity typing, and NER, whose annotations are created by prompting ChatGPT to invent appropriate labels for each sample and identify corresponding answers in an open-domain setting. The prompt is demonstrated in Section~\ref{sec:pt_gen}. Finally, the PT dataset encompasses 1,146,271 instances and 817,075 distinct labels. Detailed statistics are shown in Table~\ref{tab:pt_stat}.

\subsubsection{Negative Label Generation}

The PT data generated by ChatGPT cannot be used for training directly because of the lack of negative labels, which are labels without answers. We adopt a simple strategy: augmenting samples in the PT data with random labels sampled from the set of all labels occurred in the corresponding PT task (i.e., CLS, ET and NER). Due to the large amount of the set (as shown in Table~\ref{tab:pt_stat}), these sampled labels are likely irrelevant to the input sentence, so it is safe to assume the absence of a corresponding answer. 


\subsection{Fine-tuning Data}

\begin{figure}[tb]
    \centering
    \includegraphics[width=\linewidth]{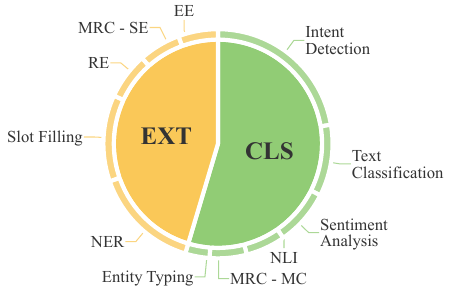}
    \caption{Ratio of each task in the fine-tuning data.}
    \label{fig:data_pie}
\end{figure}

To further calibrate models to perform NLU tasks and eliminate effects caused by errors in the PT dataset, we collect massive high-quality NLU datasets from different domains for fine-tuning. As illustrated in Figure~\ref{fig:data_pie}, our fine-tuning (FT) dataset consists of 110 NLU datasets across two languages, English and Chinese, and ten tasks, including IE tasks, such as NER, RE, and EE and other tasks which can be translated into the two atomic tasks, such as NLI and MRC. Besides a broad coverage of tasks, the data diversity is also guaranteed by their assorted source domains, including medicine, news, and dialogue with AI assistants, and different labels or queries with various granularity. Each task is translated into a combination of atomic tasks, resulting in 139 classification atomic tasks and 94 extraction atomic tasks. We manually select a small portion of the NLU datasets as the held-out set for zero-shot evaluation. A complete list of the included datasets is available in Section~\ref{sec:tasks_datasets}. 

\subsubsection{Balancing data}

A large number of datasets are collected in our FT data to ensure diversity, but meanwhile, this introduces data imbalance. Taking two classification datasets as examples, \texttt{IFLYTEK}~\cite{xu-etal-2020-clue} and \texttt{AG News}~\cite{NIPS2015_250cf8b5} contains 124 and 31,900 instances per label in average, respectively. In our implementation, we combine collected and sample data uniformly and randomly. The imbalance potentially causes underfitting tasks with abundant samples or oversampling on small datasets. Therefore, we set a quota for each dataset-label pair for balancing data. We use the whole set of instances without up-sampling for those dataset-label pair with fewer instances than the quota. 

\subsection{Two-stage Training}

We train SeqGPT based on BLOOMZ~\cite{muennighoff2023crosslingual}\footnote{Checkpoints are downloaded from the \texttt{huggingface} website: \url{https://huggingface.co/bigscience/bloomz}.}, an instruction-tuned variant of BLOOM~\cite{workshop2023bloom}, with a two-stage training strategy, including pre-training and fine-tuning, as an allusion to the usage of different training data. In our preliminary experiments, this strategy outperforms the alternative: training with a simple mixing of the PT and FT data. Specifically, we use padding to build batches and mask out training losses on the input tokens. Most hyper-parameters, including optimization steps, learning rates, and batch size, are consistent across all experiments. See Section~\ref{sec:hp} for details. 


\begin{table*}[tb]
    \centering
    \resizebox{\linewidth}{!}{
    \begin{tabular}{l|c|rrrrr rrrrrr}
    \toprule
    \textbf{Model} & \textbf{Size} & \textbf{CLS} & \textbf{EE} & \textbf{ID} & \textbf{MRC} & \textbf{NER} & \textbf{NLI} & \textbf{RE} & \textbf{SF} & \textbf{SA} & \textbf{ET} & \textbf{ALL}\\
 \midrule
       ChatGPT & - &  58.0 & 34.8 
& 62.3 &  19.9 & 11.1 & 33.5 & 31.4 & 30.6 & 65.6 & 27.9 & 38.1 \\
\midrule
       \multirow{4}{*}{BLOOMZ} & 560M  & 5.3 & 1.6 & 3.6 & 4.4 & 0.0 & 5.8 & 0.7 & 0.0 & 11.30 & 3.3 & 3.6 \\
        & 1B7 & 5.6 & 2.4 & 0.9 & 3.8 & 0.0 & 10.1 & 4.3 & 0.0 & 16.0 & 3.5 & 3.7\\
        & 3B & 6.8 & 3.9 & 1.8 & 4.4 & 0.0 & 4.4 & 3.3 & 0.0 & 12.5 & 3.6 & 4.7 \\
        & 7B1 & 10.3 & 6.2 & 2.4 & 6.4 & 0.0 & 14.0 & 11.2 & 0.2 & 24.6 & 4.2 & 6.2\\
\midrule
       & 560M  & 53.7 & 48.0 & 64.1 & 39.1 & 48.9 & 48.7 & 40.5 & 66.1 & 71.2 & 32.8 & 53.9 \\
    {SeqGPT} & 1B7 & 62.5 & 55.1 & 78.0 & 45.1 & 52.0 & 52.9 & 50.4 & 65.4 & \textbf{78.5} & 34.2 & 60.1 \\
       {w/o pre-training} & 3B & 65.9 & 59.7 & 79.9 & 45.4 & 53.8 & 57.9 & 51.6 & 70.1 & \underline{76.0} & 37.4 & 62.2 \\
       & 7B1 & \textbf{72.7} & \textbf{63.4} & \textbf{83.3} & \underline{49.2} & \underline{55.5} &  \underline{60.4} & \textbf{57.4} & 71.7 & 73.5 & 43.1 & \underline{65.4} \\
\midrule
       \multirow{4}{*}{SeqGPT} & 560M  & 57.3 & 56.8 & 72.9 & 38.8 & 50.9 & 51.4 & 43.9 & 70.0 & 71.7 & 38.8 & 57.2 \\
        & 1B7 & 67.9 & 57.2 & \underline{80.9} & 43.8 & 52.7 & 57.5 & \underline{56.7} & 70.1 & 77.2 & 48.1 & 62.8\\
       & 3B &  68.5 & 60.9	& 77.2 & 48.8 & 54.8 & \textbf{62.5} & 54.3 & \textbf{75.1} & 73.1 & \underline{48.9} & 64.0\\
       & 7B1 & \underline{70.9} & \underline{63.1} & \underline{80.9} & \textbf{51.0} & \textbf{56.1} & 58.9 & 56.0 & \underline{72.1} & 74.3 & \textbf{54.1} & \textbf{65.5} \\
       \bottomrule
    \end{tabular}
    }
    \caption{Performance on held-out evaluation datasets. CLS: text classification. EE: event extraction. ID: intent detection; MRC: machine reading comprehension. NER: named-entity recognition.  NLI: natural language inference. RE: relation extraction. SF: slot filling. SA: sentiment analysis. ET: entity typing. ALL: average performance on all tasks.}
    \label{tab:main_results}
\end{table*}

\section{Experiments}

\subsection{Evaluation}

Given the fact that LLMs sometimes generate reasonable but not exactly matched answers, the traditional Micro-F1 metric is not smooth enough for evaluation. To mitigate this and make the evaluation more minor-flaw-tolerant, we propose to combine Micro-F1 and a more smooth ROUGE score as the overall metric. Specifically, we take the average of ROUGE-1, ROUGE-2, and ROUGE-L~\cite{lin-2004-rouge}\footnote{We use the \texttt{evaluate} package to compute ROUGE scores: \url{https://github.com/huggingface/evaluate}.} as ROUGE score and take the average of Micro-F1 and ROUGE score as the final score.

To thoroughly evaluate the generalization ability, we evaluate SeqGPT on 233 held-in datasets and 49 held-out datasets. Specifically, the training split of held-in datasets is used during training, no sample from held-out datasets is seen during training, and all tasks involved in held-out datasets are seen during training. For efficiency, we randomly sample 48 records from each evaluation dataset's valid and test split.
Besides, in terms of tasks translated to multiple atomic tasks, we simplify the evaluation to report the average scores over atomic tasks. Unless otherwise specified, all scores reported in this section are held-out performance for simplicity.

\subsection{Baselines}

We compared SeqGPT with the well-known large chat language model ChatGPT \cite{openai2022chatgpt} and instruction fine-tuned model series BLOOMZ \cite{bigscience-2022-bigscience} to demonstrate the effectiveness of our method.

\subsection{Main Results}

We compared the held-out performance of the SeqGPT family and baselines in Table \ref{tab:main_results}. Based on the results, we have the following findings:

\noindent\paragraph{(1)} The smallest SeqGPT-560M surpasses the performance of ChatGPT by a large margin of 27.4, demonstrating the effectiveness of our framework and powerful natural language understanding ability can be learned by a compact small model.  On the other hand, the overall score of ChatGPT might be hindered by the metric we adopted since the output format generated by ChatGPT is not always aligned with our evaluation data format. Besides, ChatGPT sometimes can not comprehend prompts, resulting in irrelevant responses.  We refer readers to Section \ref{sec:human_eval} for a more detailed analysis of comparing ChatGPT with SeqGPT.

\noindent\paragraph{(2)} The average score can be further improved to 65.5 by using a larger 7B1 backbone. This improvement can be attributed to better complex reasoning ability and more diverse world knowledge that comes with larger pre-trained language models.

\noindent\paragraph{(3)} The weakly supervised ultra-fine-grained pre-training data are helpful, especially for smaller models. Without using the pre-training data, the performance of SeqGPT drops from 57.2 to 53.9. Specifically, the score of entity typing, which requires a diverse range of understanding of entities, drops significantly for SeqGPT of all sizes.

\noindent\paragraph{(4)} Though effective, the performance gains achieved by utilizing pre-training data shrinks with larger models. We argue that this is because the ultra-fine-grained knowledge in our pre-training data can also be learned directly during the pre-training stage of LLMs, and such knowledge is better learned with increasing model size of pre-trained LLMs. On the other hand, the naive BLOOMZ 7B1 lags far behind even the smallest SeqGPT 560M. We find the output generated by BLOOMZ 7B1 can hardly be consistent with the instruction, indicating complex prompt engineering or few-shot examples might be required to leverage such general instruction following model to solve open-domain NLU tasks.

\subsection{Scaling Analysis}

We extensively study the performance of models with respect to the scaling of model sizes, number of samples per task, and number of distinct tasks and discover all these factors are crucial for building an open-domain sequence understanding model.

\begin{figure}[tb]
    \centering
    \includegraphics[width=\linewidth]{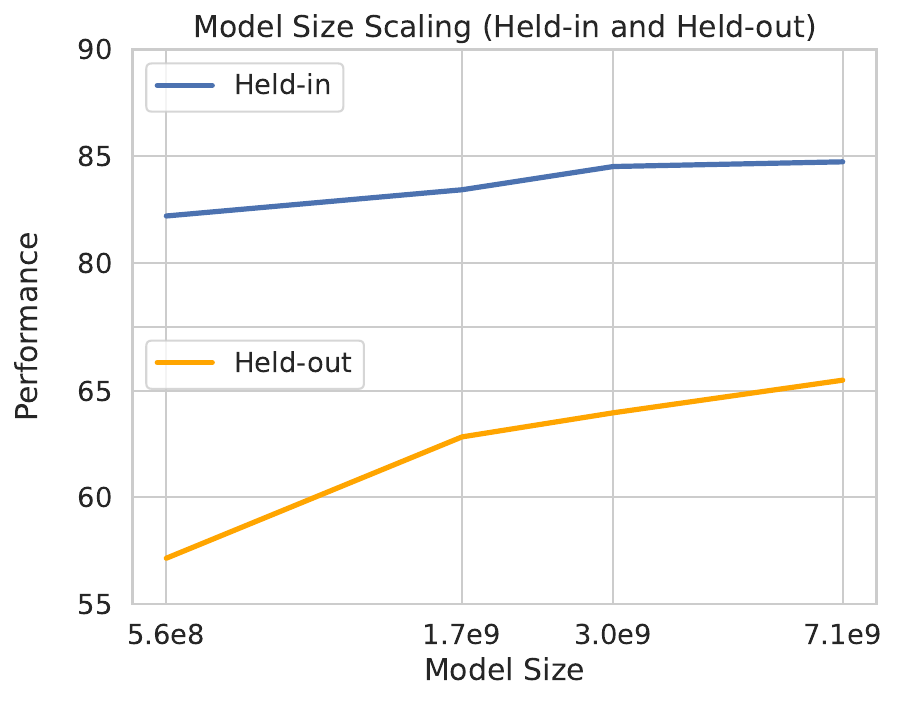}
    \caption{Held-in and held-out evaluation results of SeqGPT in different sizes.}
    \label{fig:model_size_scaling}
\end{figure}

\begin{figure}[tb]
    \centering
    \includegraphics[width=\linewidth]{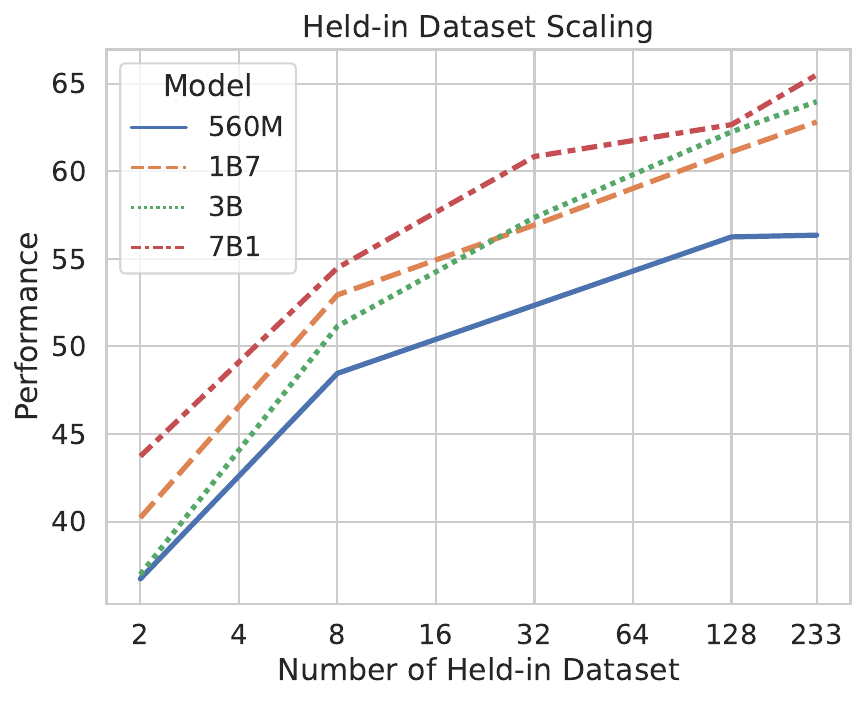}
    \caption{Held-out performance of SeqGPT in different sizes scaling with respect to the number of training datasets in the held-in set.}
    \label{fig:held-in-dataset-scaling}
\end{figure}

\begin{figure}[tb]
    \centering
    \includegraphics[width=\linewidth]{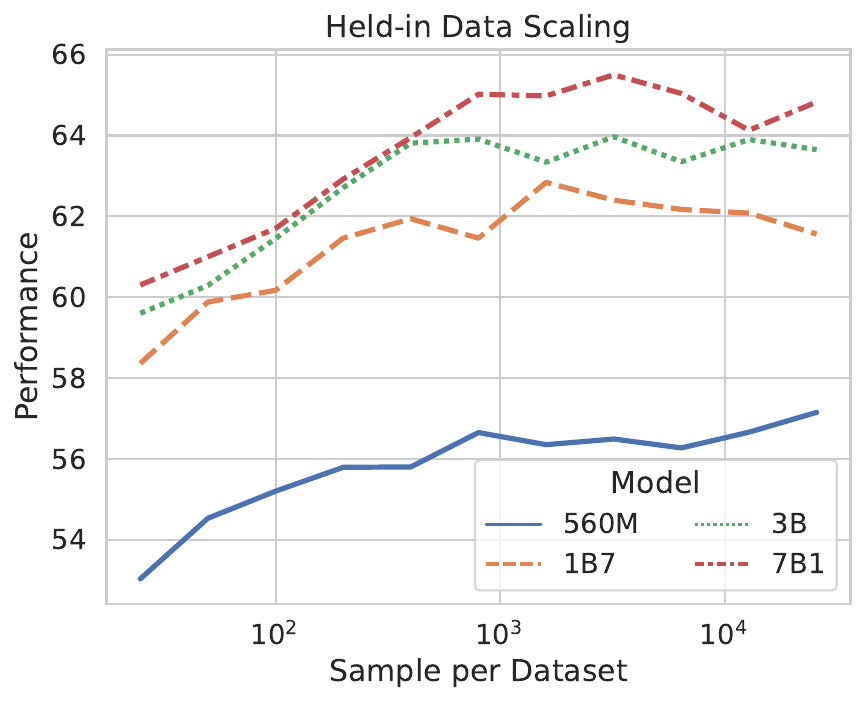}
    \caption{Held-out performance of SeqGPT scaling with respect to the number of samples per dataset.}
    \label{fig:held-in-size-scaling}
\end{figure}

\subsubsection{Model Size}

We trained a series of models in different sizes based on the BLOOMZ family \cite{bigscience-2022-bigscience} from 560M to 7B1 to explore the scaling effect of model sizes. Results in Figure \ref{fig:model_size_scaling} show both the held-in and the held-out performance increase with a larger backbone that complies with the results found in \citet{chowdhery2022palm}. Furthermore, the large gap between the held-in and held-out performance reveals the difficulty of open-domain NLU, indicating that there is still great space for SeqGPT to improve the generalization ability. We find the improvement in held-in evaluation is fewer compared with the held-out evaluation. We believe the held-out score can better reflect the performance in real applications. Besides, the performance gap between SeqGPT-7B1 and SeqGPT-3B is much smaller than the gap between SeqGPT-1B7 and SeqGPT-560M, indicating the boost of larger backbone decreases.

\subsubsection{Number of Training Datasets}

Besides the model size, the number of training datasets is also the major factor to impact the resulting performance, so we also conduct extensive experiments to explore this effect. Results in Figure \ref{fig:held-in-dataset-scaling} indicate that the performance of our SeqGPT models increases in a logarithmic manner with more datasets used for training. Based on such observation, we believe that adding more training datasets is an efficient and straightforward approach to improve the performance further since our held-in corpora are still small compared to opulent real application scenarios.

\subsection{Cross-language Generalization}

We use a great amount of training data from both English and Chinese. To explore the effect of data from each language and the cross-language generalization ability of SeqGPT, we conduct extensive experiments, and the main results are shown in Table \ref{tab:cross_language}. We can see that the models trained with a single language (English/Chinese) can generalize to tasks in the other language (Chinese/English) and achieve reasonable performance. Comparing the model trained with data in English and in both languages, we find the scores on both English tasks and Chinese tasks can be improved, showing there are skills shared between languages that can be learned through a multilingual training stage.

\begin{table}[tb]
    \centering
    \resizebox{\linewidth}{!}{
    \begin{tabular}{l|c|c}
    \toprule
    \textbf{Training Languages} & \textbf{EN Score} & \textbf{ZH Score} \\
    \midrule
     English & 57.59    & 51.98 \\
     Chinese & 52.66    & 64.57 \\
     \midrule
     Chinese + English & \textbf{58.83}    & \textbf{65.23} \\
     \bottomrule
    \end{tabular}%
    }
    \caption{Performance of SeqGPT trained with different settings of training languages. }
    \label{tab:cross_language}
\end{table}

\subsection{Cross-task Generalization}

\begin{figure}[tb]
    \centering
    \includegraphics[width=\linewidth]{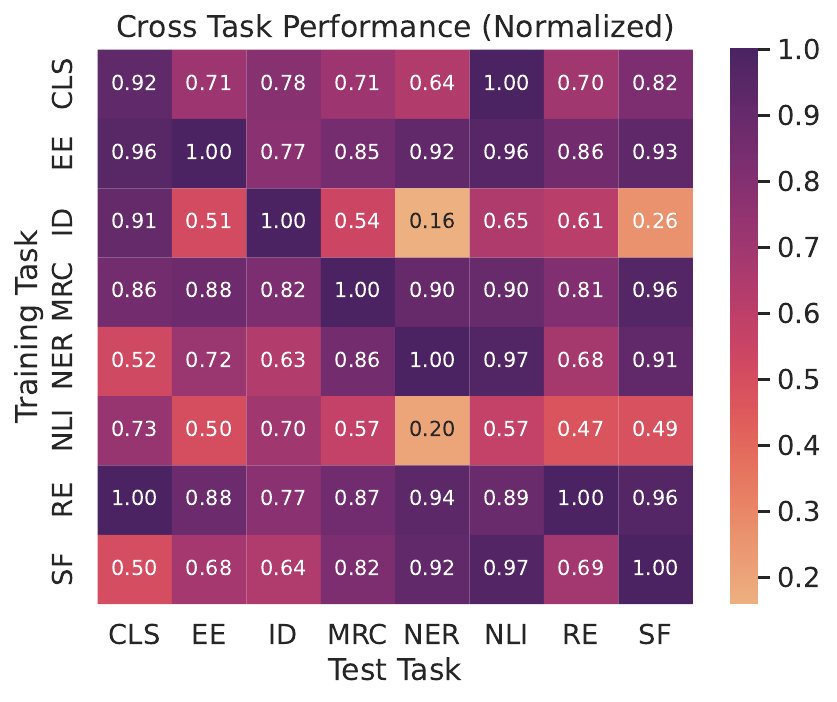}
    \caption{Cross task generalization experiment results. Scores are normalized column-wise based on the max score of each column.}
    \label{fig:cross_task}
\end{figure}

Though sharing mostly the same prompts in our framework, the skills needed to solve different tasks is diverse. To analyze how SeqGPT works on tasks not seen during training and how the training task affects the performance of different test tasks, we train a series of models with only one task, and results are shown in Figure \ref{fig:cross_task}. Based on the results we find models achieve the best evaluation performance when the evaluation task is the same as the training task except for the NLI task. For NLI performance, we find the model trained on the NLI task even achieves the worst performance. We argue this is because the way to classify sentence pairs differs across NLI datasets. As a result, models trained on only NLI datasets can hardly transfer the classification boundaries learned from the held-in datasets to held-out datasets. Models trained on EE, MRC, and RE can generalize well to all test tasks, demonstrating the diverse knowledge required to solve these tasks are also crucial for other tasks and can serve as a great training resource for models targeting general domain NLU.

\subsection{Human Evaluation}
\label{sec:human_eval}

\begin{figure}[tb]
    \centering
    \includegraphics[width=\linewidth]{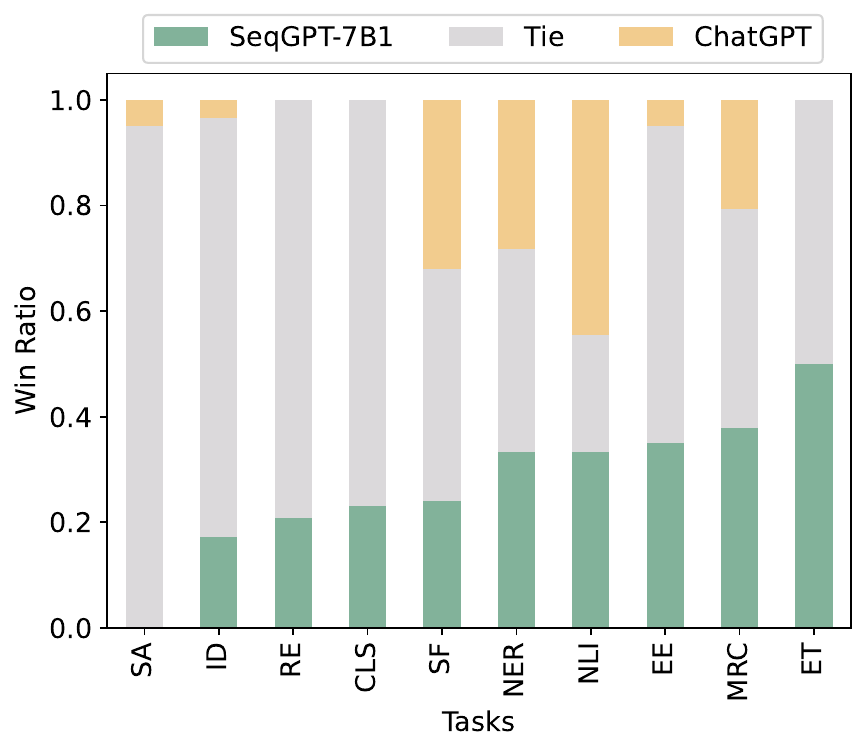}
    \caption{Human evaluation on held-out datasets.}
    \label{fig:human_eval}
\end{figure}

For a more comprehensive analysis, we perform a human evaluation on the held-out datasets. The evaluation recruits ten well-educated annotators and presents them with answers generated by ChatGPT and SeqGPT-7B1. Annotators are required to decide which model gives the better answer or two models are tied with each other. Results are shown in Figure \ref{fig:human_eval}. From the results, we can find that SeqGPT-7B1 achieves higher performance on seven out of ten NLU tasks, demonstrating the effectiveness of training the model with a wide range of NLU tasks incorporating a
great diversity of open-domain data. Also, we found the output of SeqGPT-7B1 is much more concise than the output of ChatGPT, making the interpretation easier and consequently reducing the engineering complexity to use the model to solve different downstream tasks. However, the results also indicate that medium-size models like SeqGPT-7B1 still lack the complex reasoning abilities to solve complicated tasks such as NER and SF.

\section{Related Work}


\subsection{Large language models}

Autoregressive language models have rapidly scaled up, reaching billions of parameters and trillions of training tokens. This has resulted in many emergent abilities such as few-shot learning, in-context learning, and reasoning~\cite{bubeck2023sparks,wei2022emergent}.  Examples include GPT-3~\cite{brown2020language}, PaLM~\cite{chowdhery2022palm,anil2023palm}, Chinchilla~\cite{hoffmann2022training}, Llama~\cite{touvron2023llama,touvron2023llama2}, GLM~\cite{du-etal-2022-glm,zeng2023glmb} and BLOOM~\cite{workshop2023bloom}. LLMs can be prompted to perform downstream tasks without training, such as ChatIE for IE tasks~\cite{wei2023zeroshot}, PromptNER for NER tasks~\cite{ashok2023promptner} and \citet{liu2023comprehensive} for text-to-SQL tasks.  We refer the readers to \cite{zhao2023survey,zheng2023judging,alpaca_eval} and references therein for more details.

In this study, we adopt BLOOMZ~\cite{muennighoff2023crosslingual}, a BLOOM-based instruction-tuned model, as the backbone due to its exceptional multilingual performance among publicly available models and superior generalization capabilities compared to BLOOM.

\subsection{Instruction tuning}

Instruction tuning~\cite{wei2022finetuned,wang-etal-2022-super,sanh2022multitask} is a novel finetuning paradigm that trains language models on numbers of tasks described using natural language instructions. It has shown potential benefits in aligning better with human preferences, yielding more truthful, useful, and less harmful output~\cite{ouyang2022training,lou2023prompt}. Furthermore, it has demonstrated enhanced task-specific performance~\cite{longpre2023flan,jang2023exploring,ivison-etal-2023-data} even tuning only on a single task~\cite{lee2023amr,gupta2023instruction,chen2023maybe}, as well as generalization capabilities for unseen tasks~\cite{wang-etal-2022-super,wang-etal-2023-self-instruct}. Most instruction-tuning methods leverage datasets covering some NLU tasks but with poor coverage of tasks and domains. For a specialized model, \citet{wang2023instructuie} train InstructUIE on wide IE tasks with various instructions and \citet{parmar-etal-2022-boxbart} build a biomedical LLM with a collection of biomedical datasets across multiple tasks with human-crafted instructions.

\subsection{Unified models for NLU}

Diverse NLU tasks emphasize different aspects of languages. Multitask learning has emerged as a prevalent topic, taking advantage of jointly modeling selected subsets of NLU tasks, such as enabling the use of more training data or modeling similarities between tasks~\cite[][among others]{10.1145/1390156.1390177,NIPS1995_bdb106a0,Caruana1997MultitaskL,miller-etal-2000-novel,JMLR:v8:sutton07a,Liu2016RecurrentNN,liu-etal-2019-multi,lu2022unified}. When incorporating more tasks, sequence generation models become compelling options because free texts may be the most straightforward way to encode all outputs of various NLU tasks. UIE~\cite{lu-etal-2022-unified} unify the inputs of IE tasks through a schema-based prompt mechanism and the outputs through the novel structural extraction language. Consequently, given suitable prompts, it can perform novel NLU tasks using the common semantic understanding ability learned. Subsequently, InstructUIE~\cite{wang2023instructuie} extends UIE by instruction tuning a stronger backbone model (e.g., Flan-T5 11B), showing strong zero-shot performance. USM~\cite{Lou_Lu_Dai_Jia_Lin_Han_Sun_Wu_2023} is another unified IE model based on a link prediction mechanism named semantic matching.

\section{Conclusions}

In this study, we introduce SeqGPT, a unified model devised to handle various NLU tasks by translating different NLU tasks into two common atomic tasks. In this way, SeqGPT offers a consistent input-output format, enabling it to solve unseen tasks by prompting arbitrarily varied label sets without tedious prompt engineering. To achieve strong generalization ability, we train the model using novel ultra fine-grained synthetic data and a massive collection of NLU datasets on various domains. The training is further enhanced with effective data balance and randomly sampled negative labels. Both automatic benchmarks and human evaluation on unseen tasks show that SeqGPT achieves consistent improvements over ChatGPT. In addition, we conduct comprehensive experiments to investigate behaviors of scaling, revealing a logarithmic correlation between the quantity of training tasks and model performance. We have also evaluated SeqGPT's ability to generalize across various tasks and languages. Nevertheless, our findings raise new questions. Why does the PT data fail to enhance SeqGPT-7B1, while an increase in FT data does? How to generate more high-quality NLU data to fill the data hunger of SeqGPT? We hope future research on these questions to further improve open-domain NLU models.



\bibliography{anthology,custom}
\bibliographystyle{acl_natbib}

\appendix

\section{Additional Results}
\label{sec:add_res}

\subsection{Training Hyper-parameters}
\label{sec:hp}

We list the major training hyper-parameters involved during the training stage of SeqGPT in Table~\ref{tab:training_hyper_params}.

\subsection{Inference Hyper-parameters}

We list the hyper-parameters used during inference stage in Table \ref{tab:inference_hyper_params}.

\subsection{Data Augmentation}

Given origin samples collected from different datasets and our pre-train corpora, we pre-process each sample into 
 $K$ instructions. We generate instructions by the following steps.  First, we sample at most $N_{pos}$ positive labels from the sample annotation, where $N_{pos}$ is a random number in range $[1, M_{pos}]$. Second, we uniformly sample at most $N_{neg}$ negative labels from the all labels in the corresponding dataset, where $N_{neg}$ is a random number in range $[1, M_{neg}]$. Finally, we encode the origin text and sampled labels with pre-defined templates listed in Table \ref{tab:prompts_ft}.

 In order to prevent the converge of our model harmed by in-balanced label distribution, we generate at most $N_{balance}$ instructions for each positive label. However, since the number of labels are extremely limited for SA and NLI datasets, we skip this process for these datasets.
We empirically found hyper-parameters listed in Table \ref{tab:data_aug} works well.

\section{Pre-training Data Generation}
\label{sec:pt_gen}

Table~\ref{tab:gen_pt_prompt} shows the prompt used to instruct ChatGPT to generate the pre-training data. 

\section{Qualitative Examples}
\label{sec:case_study}

Table~\ref{tab:cases} shows examples from different tasks. Each example consists of a sentence (or a phrase) and a set of label as the input, outputs from ChatGPT and SeqGPT, and the ground-truth answer. The prompt template for ChatGPT is shown in Figure~\ref{fig:intro_demo} and that for SeqGPT is shown in Table~\ref{tab:prompts_ft}.

\section{Tasks and Datasets}
\label{sec:tasks_datasets}

Prompts used in the fine-tuning tasks are listed in Table~\ref{tab:prompts_ft}. 
All public datasets in the fine-tuning dataset and the open-domain benchmark are listed in Table \ref{tab:ft_data_full}. There are also two private text classification datasets and nine private NER datasets used in the fine-tuning, which are in Chinese and from various domains, such as medicine and e-commerce.

\begin{table}[htb]
    \centering
    \resizebox{\linewidth}{!}{
    \begin{tabular}{p{3cm}|cccc}
    \toprule
    \multirow{2}{*}{\textbf{Hyper-parameter}} & \multicolumn{4}{c}{\textbf{SeqGPT}}\\
     & 560M & 1B7 & 3B & 7B1 \\\midrule
    Batch size & 4 & 4 & 2 & 1\\
    Grad accumulation & 32 & 32 &64&128\\
    Learning rate & \multicolumn{4}{c}{1e-4}\\
    Max training steps & \multicolumn{4}{c}{4000}\\
    \bottomrule
    \end{tabular}
    }
    \caption{Training Hyper-parameters.}
    \label{tab:training_hyper_params}
\end{table}

\begin{table}[htb]
    \centering
    \begin{tabular}{l|c}
    \toprule
    \textbf{Hyper-parameter} & \textbf{Value} \\\midrule
    Strategy & Beam Search \\
    Beam size & 4 \\
    Max answer tokens & 128 \\
    $\tau$ & 1.0 \\
      \bottomrule
    \end{tabular}
    \caption{Hyper-parameters used during inference.}
    \label{tab:inference_hyper_params}
\end{table}

\begin{table}[htb]
    \centering
    \begin{tabular}{cccc}
    \toprule
        $\boldsymbol{K}$ & $\boldsymbol{M_{pos}}$ & $\boldsymbol{M_{neg}}$ & $\boldsymbol{N_{balance}}$ \\
        \midrule
        3 & 11 & 21 & 500 
        \\
        \bottomrule
    \end{tabular}
    \caption{Hyper-parameters used for data augmentation}
    \label{tab:data_aug}
\end{table}

\newpage

\begin{table*}[tb]
    \centering
    \begin{CJK*}{UTF8}{gbsn}
    \begin{tabular}{p{1cm}|p{1cm}|p{10cm}}\toprule
    \textbf{Task} & \textbf{Lang} & \textbf{Prompt} \\\midrule
    CLS & En & You are asked to do the following 3 tasks: text classification, sentiment analysis, intent detection. Here are the requirements: 1. The text should be classified into at least 5 categories, separated by "/". 2. Sentiment should be in one of positive, negative or neutral. 3. The intent should contain at most 2 words describing what the text wants to do. 4. The output should be in json format. 5. Do not return the original text. "\{text\}" \\\cmidrule{2-3} 
    & Zh & 我们要对下面这句话做3个任务：文本分类、情感分析、意图识别。要求：1. 至少预测5个类别，类别之间用/分割。 2. 情感分类通常分为正向、负向和中性三类。3. 意图识别只用两个词概括，不要输出其他内容。 4. 结果使用json格式返回。“\{text\}” \\\midrule
    ET NER & En & Given the following text, identify all fine-grained entities and assign no less than three entity types to each entity. "\{text\}"\\\cmidrule{2-3} 
    & Zh & 给定下面文本，识别所有细粒度实体，并对每个实体打标不少于三个实体类型。“\{text\}”\\\bottomrule
    \end{tabular}
    \caption{Prompts used for generating the pre-training data.}
    \label{tab:gen_pt_prompt}
    \end{CJK*}
\end{table*}

\begin{table*}[tb]
    \centering
    \begin{CJK*}{UTF8}{gbsn}
    \small
    \begin{tabular}{cc|l|l}\toprule
        \multicolumn{2}{c|}{\textbf{Task}} & \textbf{Prompt} & \textbf{Translation (for references)} \\\midrule
        \multicolumn{2}{c|}{\makecell{CLS\\ID\\SA\\MRC-MC}} & \makecell[l]{输入: \{text\}\\分类: \{label\_set\}\\输出:} & \makecell[l]{Input: \{text\}\\Classify: \{label\_set\}\\Output:} \\\midrule
       \multicolumn{2}{c|}{ET} & \makecell[l]{输入: \{text\} \{mention\}\\分类: \{label\_set\}\\输出:} & \makecell[l]{Input: \{text\} \{mention\}\\Classify: \{label\_set\}\\Output:}\\\midrule
        \multicolumn{2}{c|}{NLI} & \makecell[l]{输入: \{text\_1\} \{text\_2\}\\分类: \{label\_set\}\\输出:} & \makecell[l]{Input: \{text\_1\} \{text\_2\}\\Classify: \{label\_set\}\\Output:}\\\midrule
        \multicolumn{2}{c|}{\makecell{NER\\SF\\MRC-SE}} & \makecell[l]{输入: \{text\}\\抽取: \{label\_set\}\\输出:} & \makecell[l]{Input: \{text\}\\Extract: \{label\_set\}\\Output:}\\\midrule\midrule
        \thead{\textbf{Task}} & \thead{\textbf{Atomic}\\\textbf{Task}}& \textbf{Prompt} & \textbf{Translation (for references)}  \\\midrule
        \multirow{2}{*}[-3.5ex]{EE} & CLS & \makecell[l]{输入: \{text\}中\{trigger\}是什么事件？\\分类: \{label\_set\}\\输出:} & \makecell[l]{Input: What is the event of \{trigger\} in \{text\}？\\Classify: \{label\_set\}\\Output:}\\\cmidrule{2-4} 
        & EXT & \makecell[l]{输入: \{text\}\\抽取: \{event\_list, augment\_list\}\\输出:} & \makecell[l]{Input: \{text\}\\Extract: \{event\_list, augment\_list\}\\Output:}\\
        \midrule
        \multirow{2}{*}[-4.5ex]{RE} & CLS & \makecell[l]{输入: \{text\}中\{subject\}和$\hookleftarrow$\\
        \quad\{object\}的关系是什么？\\分类: \{relation\_type\_list\}\\输出:} & \makecell[l]{Input: What is the relation between \{subject\} and $\hookleftarrow$\\
        \quad\{object\} in \{text\}?\\Classify: \{relation\_type\_list\}\\Output:}\\\cmidrule{2-4} 
        & EXT & \makecell[l]{输入: \{text\}\\抽取: \{relation\}关系的宾语，$\hookleftarrow$\\
        \quad\{relation\}关系的主语\\输出:} & \makecell[l]{Input: \{text\}\\Extract: the object of \{relation\}, the subject of $\hookleftarrow$\\
        \quad\{relation\}\\Output:}\\
        \bottomrule
    \end{tabular}
    \end{CJK*}
    \caption{Prompts used in the fine-tuning tasks}
    \label{tab:prompts_ft}
\end{table*}

\onecolumn

\newpage

\input{minipages/cases.tex}

\newpage

\input{minipages/ft_data_info}

\end{document}

%% file: minipages/cases.tex
\begin{CJK*}{UTF8}{gbsn}
\begin{longtable}{p{15cm}}
\caption{Qualitative examples\label{tab:cases}}\\
\toprule\endfirsthead
\small\sl continued from previous page\\\toprule\endhead
\bottomrule\endlastfoot
\bottomrule\small\sl continued on next page\\\endfoot
\cellcolor{gray!20}Indent detection (atomic task: classification) \\
\colorbox{highlightgray}{\textcolor{white}{Sentence:}} Let's confirm the details. You want Grocery run alarm set for 5:15 pm. \\
\colorbox{highlightgray}{\textcolor{white}{Labels:}} alarm\_time, new\_alarm\_time, new\_alarm\_name, alarm\_name \\
\colorbox{highlightgreen}{\textcolor{white}{ChatGPT:}}  alarm\_time: 5:15 pm \\
\colorbox{highlightblue}{\textcolor{white}{SeqGPT:}} new\_alarm\_name: Grocery\texttt{\textcolor{gray}{ \textbackslash n }}alarm\_time: 5:15 pm  \\
\colorbox{highlightred}{\textcolor{white}{Ground-truth:}}  new\_alarm\_time: 5:15 pm\texttt{\textcolor{gray}{ \textbackslash n }}new\_alarm\_name: Grocery run  \\\midrule 
\cellcolor{gray!20}Named entity recognition (atomic task: extraction) \\
\colorbox{highlightgray}{\textcolor{white}{Sentence:}} A frame language is a technology used for knowledge representation in artificial intelligence . \\
\colorbox{highlightgray}{\textcolor{white}{Labels:}} programlang, country, researcher, organisation, product, field, task \\
\colorbox{highlightgreen}{\textcolor{white}{ChatGPT:}}  programlang:frame language\texttt{\textcolor{gray}{ \textbackslash n }}field:knowledge representation\texttt{\textcolor{gray}{ \textbackslash n }}organisation:artificial intelligence \\
\colorbox{highlightblue}{\textcolor{white}{SeqGPT:}}  field: artificial intelligence\texttt{\textcolor{gray}{ \textbackslash n }}task: knowledge representation  \\
\colorbox{highlightred}{\textcolor{white}{Ground-truth:}}  task: knowledge representation\texttt{\textcolor{gray}{ \textbackslash n }}field: artificial intelligence  \\\midrule 
\cellcolor{gray!20}Event extraction (atomic task: extraction) \\
\colorbox{highlightgray}{\textcolor{white}{Sentence:}} I live in Redwood City, which they actually moved the trial here a couple months into it \\
\colorbox{highlightgray}{\textcolor{white}{Labels:}} Conflict/Demonstrate event, the Vehicle of event Movement/Transport, the Agent of event Life/Die, the Victim of event Life/Injure, the Place of event Justice/Appeal, the Defendant of event Justice/Sentence, the Agent of event Life/Injure, the Adjudicator of event Justice/Appeal, Justice/Trial-Hearing event, the Plaintiff of event Justice/Sue, Life/Die event, Justice/Sue event, the Place of event Justice/Trial-Hearing, the Adjudicator of event Justice/Charge-Indict, the Place of event Transaction/Transfer-Money, Life/Divorce event, the Defendant of event Justice/Charge-Indict, the Crime of event Justice/Trial-Hearing, the Place of event Life/Die \\
\colorbox{highlightgreen}{\textcolor{white}{ChatGPT:}}  Place: Redwood City\texttt{\textcolor{gray}{ \textbackslash n }}Justice/Trial-Hearing event \\
\colorbox{highlightblue}{\textcolor{white}{SeqGPT:}}  the Place of event Justice/Trial-Hearing: here\texttt{\textcolor{gray}{ \textbackslash n }}Justice/Trial-Hearing event: trial	it  \\
\colorbox{highlightred}{\textcolor{white}{Ground-truth:}}  Justice:Trial-Hearing event: trial it\texttt{\textcolor{gray}{ \textbackslash n }}the Place of event Justice:Trial-Hearing: here \\\midrule 
\cellcolor{gray!20}Event extraction (atomic task: classification) \\
\colorbox{highlightgray}{\textcolor{white}{Sentence:}} 信中说，以色列１２日对加沙和约旦河西岸进行的袭击等于是对巴 勒斯坦人民“全面宣战”，这些军事行动在巴勒斯坦被占领土上造 成了大量人员伤亡和财产损失，违反了包括联合国决议在内的国际 准则，给地区冲突带来了新的不祥之兆 中 亡是什么事件？ \\
\colorbox{highlightgray}{\textcolor{white}{Labels:}} 法律/逮捕入狱, 个人/提名, 法律/宣判无罪, 生活/死亡, 个人/选举, 移动/运输, 交易/资金流动, 商业/组织终结, 法律/控罪起诉, 法律/赦免, 生活/结婚, 法律/引渡, 法律/罚款, 法律/审讯, 冲突/示威, 商业/宣布破产 \\
\colorbox{highlightgreen}{\textcolor{white}{ChatGPT:}}  冲突/示威 \\
\colorbox{highlightblue}{\textcolor{white}{SeqGPT:}}  生活/死亡  \\
\colorbox{highlightred}{\textcolor{white}{Ground-truth:}}  生活/死亡  \\\midrule 
\cellcolor{gray!20}Named entity recognition (atomic task: extraction) \\
\colorbox{highlightgray}{\textcolor{white}{Sentence:}} They are mistaken , " said Japanese UN envoy Kenzo Oshima , whose country is bidding for permanent membership along with its G4 partners -- Brazil , Germany and India .	They \\
\colorbox{highlightgray}{\textcolor{white}{Labels:}} box\_office, artifact, nation, implement, political\_scientist, promotion, due\_process, globe, authority, datum, government \\
\colorbox{highlightgreen}{\textcolor{white}{ChatGPT:}}  政府 \\
\colorbox{highlightblue}{\textcolor{white}{SeqGPT:}}  government  \\
\colorbox{highlightred}{\textcolor{white}{Ground-truth:}}  government, authority, nation, organization  \\\midrule 
\cellcolor{gray!20}Entity typing (atomic task: extraction) \\
\colorbox{highlightgray}{\textcolor{white}{Sentence:}} 我觉得那些书对我来讲真是太肤浅了，只用一两天，就可以结束一个学期的课程。	我 \\
\colorbox{highlightgray}{\textcolor{white}{Labels:}} 作家, 法院, 电子传媒, 电影中心, 内容, 国际比赛, 小说家, 检察机关, 商品, 见证人, 电视频道, 制片公司, 人, 政策 \\
\colorbox{highlightgreen}{\textcolor{white}{ChatGPT:}}  作家、内容、小说家 \\
\colorbox{highlightblue}{\textcolor{white}{SeqGPT:}}  人  \\
\colorbox{highlightred}{\textcolor{white}{Ground-truth:}}  人  \\\midrule 
\cellcolor{gray!20}Slot filling (atomic task: extraction) \\
\colorbox{highlightgray}{\textcolor{white}{Sentence:}} 给我放白龙马 \\
\colorbox{highlightgray}{\textcolor{white}{Labels:}} 歌曲数量, 歌手名, 主题曲类型, 专辑名, 乐器, 歌曲名, 序列号, 音乐类型, 页码, 应用名, 适用年龄, 适用人群, 音乐场景, 操作, 适用人名, 主题, 音乐风格, 对象 \\
\colorbox{highlightgreen}{\textcolor{white}{ChatGPT:}}  乐器：白龙马\texttt{\textcolor{gray}{ \textbackslash n }}歌曲名：白龙马\texttt{\textcolor{gray}{ \textbackslash n }}音乐场景：白龙马 \\
\colorbox{highlightblue}{\textcolor{white}{SeqGPT:}}  歌曲名: 白龙马\texttt{\textcolor{gray}{ \textbackslash n }}操作: 放  \\
\colorbox{highlightred}{\textcolor{white}{Ground-truth:}}  操作: 放\texttt{\textcolor{gray}{ \textbackslash n }}歌曲名: 白龙马  \\\midrule 
\cellcolor{gray!20}Machine reading comprehension (atomic task: classification) \\
\colorbox{highlightgray}{\textcolor{white}{Sentence:}} the best method for detecting texture is \\
\colorbox{highlightgray}{\textcolor{white}{Labels:}} rubbing it, seeing it, hearing it, tasting it \\
\colorbox{highlightgreen}{\textcolor{white}{ChatGPT:}}  tasting it \\
\colorbox{highlightblue}{\textcolor{white}{SeqGPT:}}  rubbing it  \\
\colorbox{highlightred}{\textcolor{white}{Ground-truth:}}  (A) rubbing it  \\\midrule 
\cellcolor{gray!20}Machine reading comprehension (atomic task: extraction) \\
\colorbox{highlightgray}{\textcolor{white}{Sentence:}} 凤梨的热量并不高,是一种热量比较低的水果,一般100克凤梨含有的热量在40卡左右,凤梨的营养价值很丰富,含有丰富的微量元素钾元素,每天适当的吃一个,可以维持体内水电解质的平衡,有利尿消肿的作用,饭后吃一个凤梨,可以促进消化酶的分泌,促进消化吸收,预防便秘,凤梨中的维生素含量比较丰富,可以延缓衰老,美白皮肤。 \\
\colorbox{highlightgray}{\textcolor{white}{Labels:}} 生命缘什么台播出, 菠萝热量高吗 \\
\colorbox{highlightgreen}{\textcolor{white}{ChatGPT:}}  生命缘什么台播出：无法抽取合适答案\texttt{\textcolor{gray}{ \textbackslash n }}菠萝热量高吗：凤梨的热量并不高, 一般100克凤梨含有的热量在40卡左右。 \\
\colorbox{highlightblue}{\textcolor{white}{SeqGPT:}}  菠萝热量高吗: 并不高\texttt{\textcolor{gray}{ \textbackslash n }}生命缘什么台播出: None  \\
\colorbox{highlightred}{\textcolor{white}{Ground-truth:}}  菠萝热量高吗: 热量比较低 \\
\end{longtable}
\end{CJK*}

%% file: minipages/ft_data_info.tex
\small
\begin{center}
\tablehead{%
\multicolumn{8}{l}{\small\sl continued from previous page}\\
\toprule
\textbf{Split}    & \textbf{Task}         & \textbf{Lang.} & \textbf{Dataset}                             & \textbf{Subset}          & \textbf{AT} & \textbf{\# Inst.} & \textbf{\# Label} \\\midrule}
\tablefirsthead{%
\toprule
\textbf{Split}    & \textbf{Task}         & \textbf{Lang.} & \textbf{Dataset}                             & \textbf{Subset}          & \textbf{AT} & \textbf{\# Inst.} & \textbf{\# Label} \\\midrule}
\tabletail{%
\bottomrule
\multicolumn{8}{l}{\small\sl continued on next page}\\
}
\tablelasttail{%
\bottomrule
}%
\topcaption{All public data used in the fine-tuning stage. $+$ denotes training tasks, while $-$ denotes test tasks. \# Inst. denotes the sum of the number of instances for training/dev/test sets.} \label{tab:ft_data_full}%
\begin{mpsupertabular}{lllp{5cm}llll}
+  & EE           & En    & MAVEN~\cite{wang-etal-2020-maven}                           & -               & CLS         & 115801   & 168      \\
+  & EE           & En    & MAVEN                            & -               & EXT         & 45039    & 168      \\
+  & EE           & Zh    & DuEE~\cite{10.1007/978-3-030-60457-8_44}                             & -               & CLS         & 17495    & 74       \\
+  & EE           & Zh    & DuEE                             & -               & EXT         & 14954    & 291      \\
+  & ID   & En    & ATIS~\cite{DBLP:conf/naacl/HemphillGD90}                                & -               & CLS         & 5871     & 22       \\
+  & ID   & En    & MultiWOZ~\cite{budzianowski-etal-2018-multiwoz}                             & Hotel           & CLS         & 18390    & 2        \\
+  & ID   & En    & MultiWOZ                            & Restaurant      & CLS         & 18722    & 2        \\
+  & ID   & En    & MultiWOZ                            & Train           & CLS         & 15901    & 2        \\
+  & ID   & En    & SGD~\cite{rastogi2020towards}                                 & Banks           & CLS         & 4510     & 2        \\
+  & ID   & En    & SGD                                 & Events          & CLS         & 27653    & 3        \\
+  & ID   & En    & SGD                                 & Flights         & CLS         & 22031    & 4        \\
+  & ID   & En    & SGD                                 & Homes           & CLS         & 8277     & 3        \\
+  & ID   & En    & SGD                                 & Hotels          & CLS         & 25641    & 4        \\
+  & ID   & En    & SGD                                 & Media           & CLS         & 7911     & 3        \\
+  & ID   & En    & SGD                                 & Movies          & CLS         & 9998     & 3        \\
+  & ID   & En    & SGD                                 & Music           & CLS         & 10084    & 4        \\
+  & ID   & En    & SGD                                 & Payment         & CLS         & 1044     & 2        \\
+  & ID   & En    & SGD                                 & RentalCars      & CLS         & 17136    & 2        \\
+  & ID   & En    & SGD                                 & Restaurants     & CLS         & 21930    & 2        \\
+  & ID   & En    & SGD                                 & Services        & CLS         & 21631    & 2        \\
+  & ID   & En    & SGD                                 & Trains          & CLS         & 2240     & 2        \\
+  & ID   & En    & SGD                                 & Buses           & CLS         & 18137    & 2        \\
+  & ID   & En    & SLURP~\cite{slurp}                  & Audio           & CLS         & 387      & 5        \\
+  & ID   & En    & SLURP                               & Cooking         & CLS         & 326      & 2        \\
+  & ID   & En    & SLURP                               & Datetime        & CLS         & 578      & 4        \\
+  & ID   & En    & SLURP                               & Email           & CLS         & 1381     & 8        \\
+  & ID   & En    & SLURP                               & General         & CLS         & 963      & 6        \\
+  & ID   & En    & SLURP                               & IOT             & CLS         & 1107     & 16       \\
+  & ID   & En    & SLURP                               & Lists           & CLS         & 793      & 6        \\
+  & ID   & En    & SLURP                               & Music           & CLS         & 469      & 7        \\
+  & ID   & En    & SLURP                               & News            & CLS         & 709      & 2        \\
+  & ID   & En    & SLURP                               & Play            & CLS         & 2024     & 9        \\
+  & ID   & En    & SLURP                               & QA              & CLS         & 1685     & 8        \\
+  & ID   & En    & SLURP                               & Recommendation  & CLS         & 596      & 5        \\
+  & ID   & En    & SLURP                               & Social          & CLS         & 565      & 4        \\
+  & ID   & En    & SLURP                               & Takeaway        & CLS         & 358      & 3        \\
+  & ID   & En    & SLURP                               & Transport       & CLS         & 805      & 6        \\
+  & ID   & En    & SLURP                               & Weather         & CLS         & 855      & 2        \\
+  & ID   & En    & SNIPS~\cite{DBLP:journals/corr/abs-1805-10190}                               & -               & CLS         & 14484    & 7        \\
+  & ID   & Zh    & CrossWOZ~\cite{zhu2020crosswoz}                            & Hotel           & CLS         & 27224    & 5        \\
+  & ID   & Zh    & CrossWOZ                            & Restaurant      & CLS         & 30134    & 5        \\
+  & ID   & Zh    & CrossWOZ                            & Subway          & CLS         & 1694     & 2        \\
+  & ID   & Zh    & CrossWOZ                            & Travel          & CLS         & 29341    & 5        \\
+  & ID   & Zh    & RiSAWOZ~\cite{quan-etal-2020-risawoz}                             & Computer        & CLS         & 9677     & 7        \\
+  & ID   & Zh    & RiSAWOZ                             & Extracurricular & CLS         & 7504     & 7        \\
+  & ID   & Zh    & RiSAWOZ                             & Flight          & CLS         & 11327    & 7        \\
+  & ID   & Zh    & RiSAWOZ                             & Gzheral         & CLS         & 28818    & 7        \\
+  & ID   & Zh    & RiSAWOZ                             & Hospital        & CLS         & 6634     & 6        \\
+  & ID   & Zh    & RiSAWOZ                             & Hotel           & CLS         & 14773    & 7        \\
+  & ID   & Zh    & RiSAWOZ                             & Movie           & CLS         & 10472    & 7        \\
+  & ID   & Zh    & RiSAWOZ                             & Restaurant      & CLS         & 13048    & 7        \\
+  & ID   & Zh    & RiSAWOZ                             & Train           & CLS         & 11495    & 7        \\
+  & ID   & Zh    & RiSAWOZ                             & Travel          & CLS         & 13620    & 7        \\
+  & ID   & Zh    & RiSAWOZ                             & TVShow          & CLS         & 11031    & 7        \\
+  & ID   & Zh    & RiSAWOZ                             & Weather         & CLS         & 11252    & 6        \\
+  & ID   & Zh    & RiSAWOZ                           & Null            & CLS         & 13       & 3        \\
+  & ID   & Zh    & SMP-2020-ECDT~\cite{zhou-etal-2020-kdconv}                       & App             & CLS         & 112      & 3        \\
+  & ID   & Zh    & SMP-2020-ECDT                       & CaptialInfo     & CLS         & 110      & 5        \\
+  & ID   & Zh    & SMP-2020-ECDT                       & ChildClassics   & CLS         & 102      & 2        \\
+  & ID   & Zh    & SMP-2020-ECDT                       & ChineseZodiac   & CLS         & 110      & 5        \\
+  & ID   & Zh    & SMP-2020-ECDT                       & Cinemas         & CLS         & 100      & 4        \\
+  & ID   & Zh    & SMP-2020-ECDT                       & CityOfPro       & CLS         & 111      & 4        \\
+  & ID   & Zh    & SMP-2020-ECDT                       & Constellation   & CLS         & 109      & 5        \\
+  & ID   & Zh    & SMP-2020-ECDT                       & Contacts        & CLS         & 100      & 2        \\
+  & ID   & Zh    & SMP-2020-ECDT                       & Email           & CLS         & 125      & 5        \\
+  & ID   & Zh    & SMP-2020-ECDT                       & Epg             & CLS         & 157      & 2        \\
+  & ID   & Zh    & SMP-2020-ECDT                       & FamilyNames     & CLS         & 103      & 4        \\
+  & ID   & Zh    & SMP-2020-ECDT                       & GarbageClassify & CLS         & 141      & 6        \\
+  & ID   & Zh    & SMP-2020-ECDT                       & HistoryToday    & CLS         & 100      & 2        \\
+  & ID   & Zh    & SMP-2020-ECDT                       & Holiday         & CLS         & 97       & 2        \\
+  & ID   & Zh    & SMP-2020-ECDT                       & Home            & CLS         & 90       & 3        \\
+  & ID   & Zh    & SMP-2020-ECDT                       & IdiomsDict      & CLS         & 154      & 7        \\
+  & ID   & Zh    & SMP-2020-ECDT                       & Joke            & CLS         & 123      & 4        \\
+  & ID   & Zh    & SMP-2020-ECDT                       & Length          & CLS         & 94       & 2        \\
+  & ID   & Zh    & SMP-2020-ECDT                       & Map             & CLS         & 134      & 2        \\
+  & ID   & Zh    & SMP-2020-ECDT                       & Message         & CLS         & 145      & 3        \\
+  & ID   & Zh    & SMP-2020-ECDT                       & Music           & CLS         & 140      & 2        \\
+  & ID   & Zh    & SMP-2020-ECDT                       & New             & CLS         & 140      & 5        \\
+  & ID   & Zh    & SMP-2020-ECDT                       & PetrolPrice     & CLS         & 100      & 2        \\
+  & ID   & Zh    & SMP-2020-ECDT                       & Poetry          & CLS         & 177      & 2        \\
+  & ID   & Zh    & SMP-2020-ECDT                       & QueryCapital    & CLS         & 150      & 6        \\
+  & ID   & Zh    & SMP-2020-ECDT                       & Stock           & CLS         & 125      & 3        \\
+  & ID   & Zh    & SMP-2020-ECDT                       & Story           & CLS         & 118      & 6        \\
+  & ID   & Zh    & SMP-2020-ECDT                       & Telephone       & CLS         & 110      & 2        \\
+  & ID   & Zh    & SMP-2020-ECDT                       & Temperature     & CLS         & 97       & 2        \\
+  & ID   & Zh    & SMP-2020-ECDT                       & TimesTable      & CLS         & 84       & 4        \\
+  & ID   & Zh    & SMP-2020-ECDT                       & Tvchannel       & CLS         & 110      & 7        \\
+  & ID   & Zh    & SMP-2020-ECDT                       & VirusSearch     & CLS         & 126      & 6        \\
+  & ID   & Zh    & SMP-2020-ECDT                       & WeightScaler    & CLS         & 100      & 2        \\
+  & ID   & Zh    & SMP-2020-ECDT                       & WordFinding     & CLS         & 98       & 2        \\
+  & MRC-MC       & Zh    & DuReader 2.0 - yesno\footnote{\url{https://ai.baidu.com/broad/introduction?dataset=dureader}}                            & -               & EXT         & 52103    & 4        \\
+  & MRC-MC       & Zh    & Dureader-Yes/No~\cite{he-etal-2018-dureader}                     & -               & EXT         & 365954   & 3        \\
+  & MRC-MC       & Zh    & ReCO~\cite{DBLP:conf/aaai/WangYZXW20}                   & -               & EXT         & 290000   & 3        \\
+  & MRC-SE       & Zh    & CAIL 2019\footnote{\url{http://cail.cipsc.org.cn/task_summit.html?raceID=1&cail_tag=2019}}                           & -               & EXT         & 41287    & -1       \\
+  & MRC-SE       & Zh    & CAIL 2020\footnote{\url{http://cail.cipsc.org.cn/task_summit.html?raceID=0&cail_tag=2020}}                           & -               & EXT         & 3719     & -1       \\
+  & MRC-SE       & Zh    & DuReader 2.0 - entity~\cite{he-etal-2018-dureader}               & -               & EXT         & 149169   & 69178    \\
+  & MRC-SE       & Zh    & SQuAD-zen\footnote{\url{https://github.com/pluto-junzeng/ChineseSquad}}                           & -             & EXT         & 76449    & 63881    \\
+  & MRC-SE       & Zh    & WebQA~\cite{li2016dataset}                               & -               & EXT         & 146890   & 42165    \\
+  & NER          & En    & BC5CDR~\cite{bc5cdr}                         & Chem            & EXT         & 13938    & 1        \\
+  & NER          & En    & BC5CDR                      & Disease         & EXT         & 13938    & 1        \\
+  & NER          & En    & BC2GM~\cite{bc2gm}                               & -               & EXT         & 20131    & 1        \\
+  & NER          & En    & BC4chemd~\cite{Krallinger2015TheCC}                            & -               & EXT         & 87685    & 1        \\
+  & NER          & En    & JNLPBA~\cite{collier-kim-2004-introduction}                              & -               & EXT         & 24806    & 5        \\
+  & NER          & En    & NCBI-disease~\cite{10.5555/2772763.2772800}                                & -         & EXT         & 7287     & 1        \\
+  & NER          & En    & anlp-sciner\footnote{\url{https://github.com/neubig/nlp-from-scratch-assignment-2022}}                 & -          & EXT         & 3978     & 15       \\
+  & NER          & En    & aspectemo~\cite{11321/849}                        & -               & EXT         & 1465     & 6        \\
+  & NER          & En    & bionlp2004~\cite{collier-kim-2004-introduction}                          & -               & EXT         & 20475    & 5        \\
+  & NER          & En    & conll03~\cite{DBLP:conf/conll/SangM03}                             & -               & EXT         & 20744    & 4        \\
+  & NER          & En    & crossner~\cite{Liu2020CrossNEREC}                    & Music               & EXT         & 945      & 13       \\
+  & NER          & En    & crossner                  & Politics               & EXT         & 1392     & 9        \\
+  & NER          & En    & crossner                   & Science               & EXT         & 1193     & 17       \\
+  & NER          & En    & fabner~\cite{Kumar2021FabNERIE}                              & -               & EXT         & 13682    & 12       \\
+  & NER          & En    & fewnerd~\cite{ding-etal-2021-nerd}                            & -               & EXT         & 188239   & 67       \\
+  & NER          & En    & multiconer22~\cite{malmasi-etal-2022-multiconer}                        & -               & EXT         & 233918   & 6        \\
+  & NER          & En    & multiconer23~\cite{fetahu-etal-2023-semeval}                        & -               & EXT         & 267629   & 33       \\
+  & NER          & En    & multinerd~\cite{tedeschi-navigli-2022-multinerd}                           & -               & EXT         & 164144   & 17       \\
+  & NER          & En    & nlpcc2022~\cite{10.1007/978-3-031-17189-5_30}                           & -               & EXT         & 223348   & 24       \\
+  & NER          & En    & ontonotes5~\cite{Weischedel2017OntoNotesA}                          & -               & EXT         & 76714    & 18       \\
+  & NER          & En    & political-advertising-pl~\cite{augustyniak-etal-2020-political}            & -              & EXT         & 1701     & 19       \\
+  & NER          & En    & re3d\footnote{\url{https://github.com/dstl/re3d}}                                & -               & EXT         & 965      & 10       \\
+  & NER          & En    & skill\_extraction~\cite{Green2022}                   & -               & EXT         & 9970     & 5        \\
+  & NER          & En    & wikidiverse~\cite{wang-etal-2022-wikidiverse}                         & -               & EXT         & 7824     & 13       \\
+  & NER          & En    & wikineural~\cite{tedeschi-etal-2021-wikineural-combined}                          & -               & EXT         & 101305   & 16       \\
+  & NER          & En    & wnut16~\cite{strauss-etal-2016-results}                              & -               & EXT         & 7244     & 10       \\
+  & NER          & En    & wnut17~\cite{derczynski2017results}                              & -               & EXT         & 5690     & 6        \\
+  & NER          & Zh    & ccks2020~\cite{10.1162/dint_a_00093}                            & -               & EXT         & 80000    & 22       \\
+  & NER          & Zh    & ccks\_medical\footnote{\url{https://www.osredm.com/competition/zstp2022/}}                       & -               & EXT         & 8864     & 6        \\
+  & NER          & Zh    & ccks\_military\footnote{\url{https://www.biendata.xyz/competition/ccks_2019_1/}}                      & -               & EXT         & 1326     & 4        \\
+  & NER          & Zh    & cluener~\cite{Xu2020CLUENER2020FN}                             & -               & EXT         & 12091    & 10       \\
+  & NER          & Zh    & datafound\_manufact\_inductry  & -               & EXT         & 1491     & 3        \\
+  & NER          & Zh    & financial\_2022                     & -               & EXT         & 11       & 4        \\
+  & NER          & Zh    & insurance\_2022                     & -               & EXT         & 30       & 7        \\
+  & NER          & Zh    & msra~\cite{levow-2006-third}                                & -               & EXT         & 45000    & 3        \\
+  & NER          & Zh    & multiconer22~\cite{malmasi-etal-2022-multiconer}                        & -               & EXT         & 167761   & 6        \\
+  & NER          & Zh    & multiconer23~\cite{fetahu-etal-2023-semeval}                        & -               & EXT         & 30530    & 33       \\
+  & NER          & Zh    & resume~\cite{zhang-yang-2018-chinese}                              & -               & EXT         & 4761     & 8        \\
+  & NER          & Zh    & zh-ontonotes~\cite{ws-2011-natural-language}                        & -               & EXT         & 24373    & 4        \\
+  & NLI          & En    & DocNLI~\cite{yin2021docnli}                           & -               & CLS         & 1443658  & 2        \\
+  & NLI          & En    & Hans~\cite{mccoy2020right}                             & -               & CLS         & 60000    & 2        \\
+  & NLI          & En    & MNLI~\cite{wang2018glue}                            & -               & CLS         & 412349   & 3        \\
+  & NLI          & En    & SNLI~\cite{bowman2015large}                             & -               & CLS         & 569033   & 3        \\
+  & NLI          & Zh    & CNSD-MNLI~\cite{xu-etal-2020-clue}                        & -               & CLS         & 410251   & 3        \\
+  & NLI          & Zh    & CNSD-SNLI~\cite{xu-etal-2020-clue}                       & -               & CLS         & 564349   & 3        \\
+  & RE           & En    & FewRel wiki~\cite{chen-li-2021-zs}                      & -               & CLS         & 67200    & 80       \\
+  & RE           & En    & FewRel wiki                      & -               & EXT         & 67200    & 160      \\
+  & RE           & En    & Semeval~\cite{gabor-etal-2018-semeval}                          & -               & CLS         & 8853     & 9        \\
+  & RE           & En    & Semeval                         & -               & EXT         & 8853     & 18       \\
+  & RE           & Zh    & DuIE~\cite{10.1007/978-3-030-32236-6_72}                             & -               & CLS         & 348534   & 48       \\
+  & RE           & Zh    & DuIE                             & -               & EXT         & 212641   & 96       \\
+  & SA           & En    & Amazon Review Full~\cite{McAuley2013HiddenFA}                  & -               & CLS         & 3650000  & 5        \\
+  & SA           & En    & Amazon Review Polarity~\cite{NIPS2015_250cf8b5}              & -               & CLS         & 4000000  & 2        \\
+  & SA           & En    & IMDB~\cite{maas-EtAl:2011:ACL-HLT2011}                                & -               & CLS         & 50000    & 2        \\
+  & SA           & En    & Yelp Review Full~\cite{NIPS2015_250cf8b5}                    & -               & CLS         & 700000   & 5        \\
+  & SA           & En    & Yelp Review Polarity~\cite{NIPS2015_250cf8b5}                & -               & CLS         & 598000   & 2        \\
+  & SA           & Zh    & CFET coarse 9~\cite{lee-etal-2020-chinese}                       & -               & CLS         & 4798     & 10       \\
+  & SA           & Zh    & \begin{CJK*}{UTF8}{gbsn}微博情感二分类\end{CJK*} (Weibo Sentiment Analysis - 2 classes)\footnote{\url{https://github.com/SophonPlus/ChineseNlpCorpus}}                       & -               & CLS         & 119988   & 2        \\
+  & SA           & Zh    & \begin{CJK*}{UTF8}{gbsn}微博情感四分类\end{CJK*} (Weibo Sentiment Analysis - 4 classes)\footnote{\url{https://github.com/SophonPlus/ChineseNlpCorpus}}                       & -               & CLS         & 361744   & 4        \\
+  & SA           & Zh    & \begin{CJK*}{UTF8}{gbsn}亚马逊商品评论情感分类数据集\end{CJK*}(Amazon Product Review)                       & -               & CLS         & 7202920  & 6        \\
+  & SA           & Zh    & \begin{CJK*}{UTF8}{gbsn}商品评论情感分类数据集\end{CJK*}(Product Review)                     & -               & CLS         & 62774    & 2        \\
+  & SA           & Zh    & \begin{CJK*}{UTF8}{gbsn}大众点评分类数据集\end{CJK*}(Dazhong Dianping)                          & -               & CLS         & 3293878  & 5        \\
+  & SA           & Zh    & \begin{CJK*}{UTF8}{gbsn}电影评论情感分类数据集\end{CJK*}(Movie Review)                       & -               & CLS         & 2125056  & 5        \\
+  & SA           & Zh    & \begin{CJK*}{UTF8}{gbsn}财经新闻情感分类数据集\end{CJK*}(Financial News)                     & -               & CLS         & 16136    & 2        \\
+  & SF & En    & ATIS~\cite{DBLP:conf/naacl/HemphillGD90}                                & -               & EXT         & 5871     & 75       \\
+  & SF & En    & MultiWOZ~\cite{budzianowski-etal-2018-multiwoz}                            & Attraction      & EXT         & 72797    & 1        \\
+  & SF & En    & MultiWOZ                            & Bus             & EXT         & 71522    & 2        \\
+  & SF & En    & MultiWOZ                            & Hospital        & EXT         & 71528    & 1        \\
+  & SF & En    & MultiWOZ                            & Hotel           & EXT         & 74004    & 3        \\
+  & SF & En    & MultiWOZ                            & Restaurant      & EXT         & 74252    & 3        \\
+  & SF & En    & MultiWOZ                            & Taxi            & EXT         & 72265    & 5        \\
+  & SF & En    & MultiWOZ                            & Train           & EXT         & 71748    & 2        \\
+  & SF & En    & SGD~\cite{rastogi2020towards}                               & Banks           & EXT         & 9635     & 7        \\
+  & SF & En    & SGD                                 & Buses           & EXT         & 36991    & 16       \\
+  & SF & En    & SGD                                 & Events          & EXT         & 58254    & 12       \\
+  & SF & En    & SGD                                 & Flights         & EXT         & 45417    & 16       \\
+  & SF & En    & SGD                                 & Homes           & EXT         & 17193    & 7        \\
+  & SF & En    & SGD                                 & Hotels          & EXT         & 55072    & 17       \\
+  & SF & En    & SGD                                 & Media           & EXT         & 17113    & 8        \\
+  & SF & En    & SGD                                 & Messaging       & EXT         & 2425     & 2        \\
+  & SF & En    & SGD                                 & Movies          & EXT         & 21240    & 16       \\
+  & SF & En    & SGD                                 & Music           & EXT         & 21339    & 5        \\
+  & SF & En    & SGD                                 & Payment         & EXT         & 2038     & 2        \\
+  & SF & En    & SGD                                 & RentalCars      & EXT         & 35163    & 11       \\
+  & SF & En    & SGD                                 & Restaurants     & EXT         & 45966    & 11       \\
+  & SF & En    & SGD                                 & RideSharing     & EXT         & 21697    & 4        \\
+  & SF & En    & SGD                                 & Services        & EXT         & 44400    & 11       \\
+  & SF & En    & SGD                                 & Trains          & EXT         & 4674     & 7        \\
+  & SF & En    & SGD                                 & Travel          & EXT         & 17462    & 3        \\
+  & SF & En    & SGD                                 & Weather         & EXT         & 9424     & 6        \\
+  & SF & En    & SNIPS~\cite{DBLP:journals/corr/abs-1805-10190}                              & -               & EXT         & 14484    & 39       \\
+  & SF & En    & movie-complex\footnote{\url{https://groups.csail.mit.edu/sls/downloads/movie/}}                       & -               & EXT         & 3906     & 12       \\
+  & SF & En    & movie-simple                        & -               & EXT         & 12218    & 12       \\
+  & SF & Zh    & CATSLU~\cite{Zhu2019}                        & Map             & EXT         & 5825     & 15       \\
+  & SF & Zh    & CATSLU                      & Video           & EXT         & 1649     & 27       \\
+  & SF & Zh    & RiSAWOZ~\cite{quan-etal-2020-risawoz}                             & -               & EXT         & 151882   & 113      \\
+  & CLS     & En    & AG News~\cite{NIPS2015_250cf8b5}                             & -               & CLS         & 127600   & 4        \\
+  & CLS     & En    & DBpedia~\cite{NIPS2015_250cf8b5}                             & -               & CLS         & 630000   & 14       \\
+  & CLS     & En    & Yahoo Answers~\cite{NIPS2015_250cf8b5}                       & -               & CLS         & 1460000  & 10       \\
+  & CLS     & En    & clinc\_full~\cite{larson-etal-2019-evaluation}                         & -               & CLS         & 23700    & 151      \\
+  & CLS     & Zh    & DuEE~\cite{Li2020DuEEAL}                   & -               & CLS         & 13456    & 65       \\
+  & CLS     & Zh    & CAIL2018~\cite{xiao2018cail2018}           & -               & CLS         & 1927870  & 202      \\
+  & CLS     & Zh    & CAIL2019\footnote{\url{http://cail.cipsc.org.cn/task_summit.html?raceID=1&cail_tag=2019}}                     & Loan               & CLS         & 8659     & 20       \\
+  & CLS     & Zh    & CAIL2019                  & Labor arbitration               & CLS         & 8513     & 20       \\
+  & CLS     & Zh    & CAIL2019                   & Marriage               & CLS         & 16115    & 20       \\
+  & CLS     & Zh    & IFLYTEK~\cite{xu-etal-2020-clue}               & -               & CLS         & 14732    & 119      \\
+  & CLS     & Zh    & Amazon Review Rating~\cite{10.1145/2736277.2741087}                        & -               & CLS         & 525619   & 1175     \\                         & -               & CLS         & 1215     & 135      \\
+  & CLS     & Zh    & Fudan News\footnote{\url{http://www.nlpir.org/wordpress/download/tc-corpus-answer.rar}}                          & -               & CLS         & 19635    & 20       \\
+  & CLS     & Zh    & TNEWS Multilevel~\cite{tnews_mlc}                  & -               & CLS         & 43761    & 1067     \\
+  & CLS     & Zh    & TNEWS~\cite{xu-etal-2020-clue}                      & -               & CLS         & 63360    & 15       \\
+  & CLS     & Zh    & \begin{CJK*}{UTF8}{gbsn}学生评语分类数据集\end{CJK*}(Student Comments)                            & -               & CLS         & 22118    & 6        \\
+  & CLS     & Zh    & \begin{CJK*}{UTF8}{gbsn}百科问答分类数据集\end{CJK*}(Wiki QA)~\cite{CLUECorpus2020}                           & -               & CLS         & 1470142  & 388      \\
+  & CLS     & Zh    & \begin{CJK*}{UTF8}{gbsn}社区问答\end{CJK*}(Forum QA)~\cite{CLUECorpus2020}                  & -               & CLS         & 4258310  & 27845    \\
+  & CLS     & Zh    & \begin{CJK*}{UTF8}{gbsn}网页层次分类数据集\end{CJK*}(Webpage Classification)\footnote{\url{https://csri.scu.edu.cn/info/1012/2827.htm}}                           & -               & CLS         & 65592    & 41       \\
- & EE           & En    & ACE05~\cite{walker_christopher_ace_2006}                       & -               & EXT         & 3577     & 157      \\
- & EE           & En    & ACE05                         & -               & CLS         & 4798     & 33       \\
- & EE           & Zh    & ACE05                         & -               & CLS         & 3164     & 33       \\
- & EE           & Zh    & ACE05                         & -               & EXT         & 2059     & 156      \\
- & ID   & En    & SGD~\cite{rastogi2020towards}                                 & Calendar        & CLS         & 5386     & 3        \\
- & ID   & En    & SGD                         & Alarm           & CLS         & 1200     & 2        \\
- & ID   & En    & SLURP~\cite{slurp}                       & Alarm           & CLS         & 550      & 4        \\
- & ID   & En    & SLURP                    & Calendar        & CLS         & 2370     & 6        \\
- & ID   & Zh    & CrossWOZ~\cite{zhu2020crosswoz}                       & Taxi            & CLS         & 1782     & 2        \\
- & ID   & Zh    & RiSAWOZ~\cite{quan-etal-2020-risawoz}                        & car             & CLS         & 5503     & 6        \\
- & ID   & Zh    & SMP-2019-NLU\footnote{\url{https://adamszq.github.io/smp2019ecdt_task1/}}                        & -             & CLS         & 2579     & 24       \\
- & MRC-MC       & En    & OpenBookQA~\cite{mihaylov-etal-2018-suit}                          & -               & EXT         & 5957     & 4        \\
- & MRC-MC       & En    & WikiQA~\cite{yang-etal-2015-wikiqa}                              & -               & EXT         & 29258    & 2        \\
- & MRC-MC       & Zh    & C3~\cite{sun2020investigating}                                  & -               & EXT         & 19102    & -       \\
- & MRC-MC       & Zh    & CAIL 2021\footnote{\url{http://cail.cipsc.org.cn/task_summit.html?raceID=0&cail_tag=2021}}                           & -               & EXT         & 25126    & -       \\
- & MRC-SE       & En    & BiPaR - en~\cite{jing2019bipar}                          & -               & EXT         & 14668    & -       \\
- & MRC-SE       & En    & SubjQA~\cite{bjerva2020subjqa}                              & -               & EXT         & 11517    & -       \\
- & MRC-SE       & Zh    & BiPaR - cn~\cite{jing2019bipar}                          & -               & EXT         & 14668    & -       \\
- & MRC-SE       & Zh    & DuReader checklist~\cite{he-etal-2018-dureader}                  & -               & EXT         & 1941     & 1924     \\
- & NER          & En    & biomedical\_anatomical\_ner~\cite{xu2014anatomical} & -               & EXT         & 4697     & 11       \\
- & NER          & En    & crossner~\cite{Liu2020CrossNEREC}                        & AI               & EXT         & 881      & 14       \\
- & NER          & En    & crossner~\cite{Liu2020CrossNEREC}                & Literature               & EXT         & 916      & 12       \\
- & NER          & En    & gum~\cite{NEURIPS2022_890b206e}                                 & -               & EXT         & 3495     & 11       \\
- & NER          & En    & legal\_ner                          & -               & EXT         & 12069    & 14       \\
- & NER          & Zh    & mmc\_diabetes\_2018                 & -               & EXT         & 3498     & 18       \\
- & NER          & Zh    & wanchuang\_medical                  & -               & EXT         & 1255     & 13       \\
- & NER          & Zh    & weibo~\cite{peng-dredze-2015-named}                               & -               & EXT         & 1889     & 8        \\
- & NLI          & En    & QNLI~\cite{wang2018glue}                             & -               & CLS         & 110206   & 2        \\
- & NLI          & Zh    & OCNLI~\cite{xu-etal-2020-clue}                               & -               & CLS         & 53387    & 3        \\
- & RE           & En    & nyt\footnote{\url{https://drive.google.com/file/d/10f24s9gM7NdyO3z5OqQxJgYud4NnCJg3/view}}                              & -               & CLS         & 2502     & 25       \\
- & RE           & En    & nyt                             & -               & EXT         & 2502     & 50       \\
- & RE           & En    & pubmed                           & -               & CLS         & 1002     & 10       \\
- & RE           & En    & pubmed                           & -               & EXT         & 1002     & 20       \\
- & RE           & Zh    & IPRE~\cite{wang2019ipre}                             & -               & CLS         & 32852    & 19       \\
- & SA           & En    & SST-2~\cite{socher-etal-2013-recursive}                               & -               & CLS         & 9613     & 2        \\
- & SA           & En    & SST-5~\cite{socher-etal-2013-recursive}                               & -               & CLS         & 11855    & 5        \\
- & SA           & Zh    & \begin{CJK*}{UTF8}{gbsn}ChnSentiCorp 酒店评论情感分类数据集\end{CJK*} (Hotel Reviews)\footnote{\url{https://github.com/pengming617/bert_classification}}            & -               & CLS         & 7765     & 2        \\
- & SA           & Zh    & \begin{CJK*}{UTF8}{gbsn}外卖评论\end{CJK*} Takeout Reviews                        & -               & CLS         & 11987    & 2        \\
- & SF & En    & SGD~\cite{rastogi2020towards}                                 & Alarm           & EXT         & 2685     & 4        \\
- & SF & En    & SGD                                 & Calendar        & EXT         & 11425    & 6        \\
- & SF & En    & MIT Restaurant~\cite{ushio-camacho-collados-2021-ner}                          & -               & EXT         & 9181     & 8        \\
- & SF & Zh    & CATSLU~\cite{Zhu2019}                      & music           & EXT         & 2224     & 19       \\
- & SF & Zh    & CATSLU                    & weather         & EXT         & 2090     & 10       \\
- & CLS     & En    & TREC~\cite{li-roth-2002-learning}                                & -               & CLS         & 5952     & 50       \\
- & CLS     & En    & BANKING~\cite{casanueva-etal-2020-efficient}                             & -               & CLS         & 13083    & 77       \\
- & CLS     & En    & StackOverflow~\cite{xu-etal-2015-short}                       & -               & CLS         & 20000    & 20       \\
- & CLS     & Zh    & CAIL 2022 Event Detection~\cite{yao-etal-2022-leven}                       & -               & CLS         & 8116     & 118      \\
- & CLS     & Zh    & THUCNews~\cite{maosong_thuctc:_2016}                    & -               & CLS         & 7000     & 14       \\
- & CLS     & Zh    & CMID~\cite{Chen2020}                       & -               & CLS         & 12254    & 36       \\
- & Typing       & En    & UFET~\cite{choi-etal-2018-ultra}                                & -               & CLS         & 5994     & 2519     \\
- & Typing       & Zh    & CFET~\cite{lee-etal-2020-chinese}                             & -               & CLS         & 4798     & 1302   \\ 
\end{mpsupertabular}
\end{center}